\documentclass[sigconf]{acmart}
\usepackage{subcaption}
\usepackage{graphicx}
\usepackage{colortbl,xcolor}
\usepackage{multirow}
\usepackage{float}

\AtBeginDocument{%
  }

\copyrightyear{2025}
\acmYear{2025}
\setcopyright{rightsretained}
\acmConference[CHI EA '25]{Extended Abstracts of the CHI Conference on Human Factors in Computing Systems}{April 26-May 1, 2025}{Yokohama, Japan}
\acmBooktitle{Extended Abstracts of the CHI Conference on Human Factors in Computing Systems (CHI EA '25), April 26-May 1, 2025, Yokohama, Japan}\acmDOI{10.1145/3706599.3720179}
\acmISBN{979-8-4007-1395-8/2025/04}

\begin{document}

\title[Does the Appearance of Autonomous Conversational Robots Affect User Spoken Behaviors in Real-World Interactions?]{Does the Appearance of Autonomous Conversational Robots Affect User Spoken Behaviors in Real-World Conference Interactions?}

\author{Zi Haur Pang}
\affiliation{%
  \institution{Kyoto University}
  \city{Kyoto}
  \country{Japan}
}
\email{pang@sap.ist.i.kyoto-u.ac.jp}

\author{Yahui Fu}
\affiliation{%
  \institution{Kyoto University}
  \city{Kyoto}
  \country{Japan}
}
\email{fu@sap.ist.i.kyoto-u.ac.jp}

\author{Divesh Lala}
\affiliation{%
  \institution{Kyoto University}
  \city{Kyoto}
  \country{Japan}
}
\email{lala@sap.ist.i.kyoto-u.ac.jp}

\author{Mikey Elmers}
\affiliation{%
  \institution{Kyoto University}
  \city{Kyoto}
  \country{Japan}
}
\email{elmers@sap.ist.i.kyoto-u.ac.jp}

\author{Koji Inoue}
\affiliation{%
  \institution{Kyoto University}
  \city{Kyoto}
  \country{Japan}
}
\email{inoue@sap.ist.i.kyoto-u.ac.jp}

\author{Tatsuya Kawahara}
\affiliation{%
  \institution{Kyoto University}
  \city{Kyoto}
  \country{Japan}
}
\email{kawahara@sap.ist.i.kyoto-u.ac.jp}


\begin{abstract}

We investigate the impact of robot appearance on users' spoken behavior during real-world interactions by comparing a human-like android, ERICA, with a less anthropomorphic humanoid, TELECO. Analyzing data from 42 participants at SIGDIAL 2024, we extracted linguistic features such as disfluencies and syntactic complexity from conversation transcripts. The results showed moderate effect sizes, suggesting that participants produced fewer disfluencies and employed more complex syntax when interacting with ERICA. Further analysis involving training classification models like Naïve Bayes, which achieved an F1-score of 71.60\%, and conducting feature importance analysis, highlighted the significant role of disfluencies and syntactic complexity in interactions with robots of varying human-like appearances. Discussing these findings within the frameworks of cognitive load and Communication Accommodation Theory, we conclude that designing robots to elicit more structured and fluent user speech can enhance their communicative alignment with humans.

\end{abstract}





\maketitle

\section{Introduction}

One of the key objectives of conversational robots is to achieve human-level interaction. To achieve this, previous research has explored various aspects, including reasoning \cite{chen2021reasoning, wang2021empathetic}, empathy \cite{pang2024acknowledgment, fu2023dual}, and personality \cite{wu2021personalized, liang2021infusing}, to enable robots to produce more human-like responses. Additionally, non-verbal interaction features, such as backchanneling \cite{inoue2024yeah, adiba2021towards}, head nodding \cite{zhou2022responsive, bayramouglu2021engagement}, and gestures \cite{nagy2021framework, ali2020automatic, trovato2016hugging}, have been extensively studied to enhance the naturalness of human-robot interactions.

Besides these communication elements, the appearance of robots remains a critical factor in shaping user perceptions and behavior. A robot's appearance strongly influences the user's first impression, particularly regarding its perceived level of human-likeness. Previous studies have shown that robots with more human-like appearances can enhance perceived warmth \cite{liu2022friendly, kim2019eliza}, empathy \cite{zlotowski2016appearance, yang2024can}, and social presence \cite{straub2016looks, sasser2024investigation}, demonstrating their effectiveness in various social and cultural settings \cite{prakash2015some, walters2008avoiding, zhang2023effects, li2010cross}.

However, research on the effect of robot appearance on user behavior faces several limitations. First, many studies involve participants interacting with images or videos of robots rather than real, physical robots, which may overlook the impact of social presence during interactions \cite{jung2023social, saeki2024impact, belanche2021examining}. Second, experiments often rely on teleoperated or Wizard-of-Oz (WoZ) methodologies, where the robot's behavior is controlled by a human operator \cite{straub2016looks, song2022costume, villani2024does}. Such setups may not fully reflect the dynamics of interactions in autonomous systems. Third, most studies are conducted in laboratory settings, where participants are recruited for controlled experiments \cite{bennett2017differences, tung2016child, rizvanouglu2014impact}. These conditions may differ significantly from real-world scenarios, potentially affecting user behavior. Finally, prior research often relies on simple metrics (e.g., conversation length, informativeness, etc.) through self-reported scales (e.g., 7-point Likert scale \cite{likert1932technique}) \cite{kanda2008analysis, zlotowski2016appearance, zhu2023key, niculescu2013making}, leaving space for deeper analysis through other perspectives.

To address these gaps, we investigate how robot appearance affect user spoken behavior during real interactions with physical, fully autonomous conversational robots at an international conference. This setting allows us to capture user responses that closely resemble everyday behavior, providing insights into how appearance influences user speech in genuine social contexts. We leverage various features in natural language processing (NLP) and conversation analysis, derived from a linguistic perspective, to offer a fine-grained analysis of user behavior. Furthermore, we developed a machine learning model capable of predicting the robot's human-likeness based on observed user behavior.

Our contributions are twofold:
\begin{itemize}
    \item We investigated how the human-likeness of an autonomous conversational robot influenced user spoken behavior in real-world interactions, from linguistic perspective.
    \item We developed a predictive model that classifies a robot's human-likeness based upon a user's spoken behavior, illuminating important linguistic cues through feature importance analysis.
\end{itemize}


\section{Dataset}

\subsection{Conversational Robots}
\begin{figure}
  \centering
  \includegraphics[width=\linewidth]{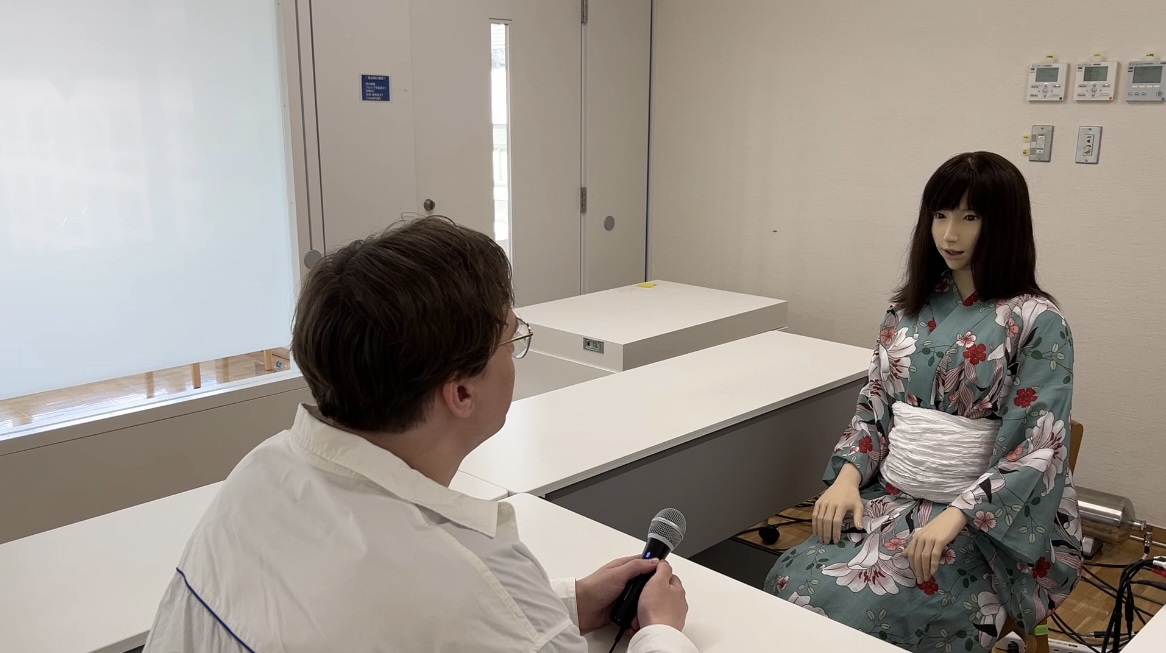} 
  \Description{Photo of interview dialogue with ERICA by SIGDIAL participant}
  \caption{
  Photo of interview dialogue with ERICA by SIGDIAL participant
  }
  \label{fig:erica_photo}
\end{figure}

\begin{figure}
  \centering
  \includegraphics[width=\linewidth]{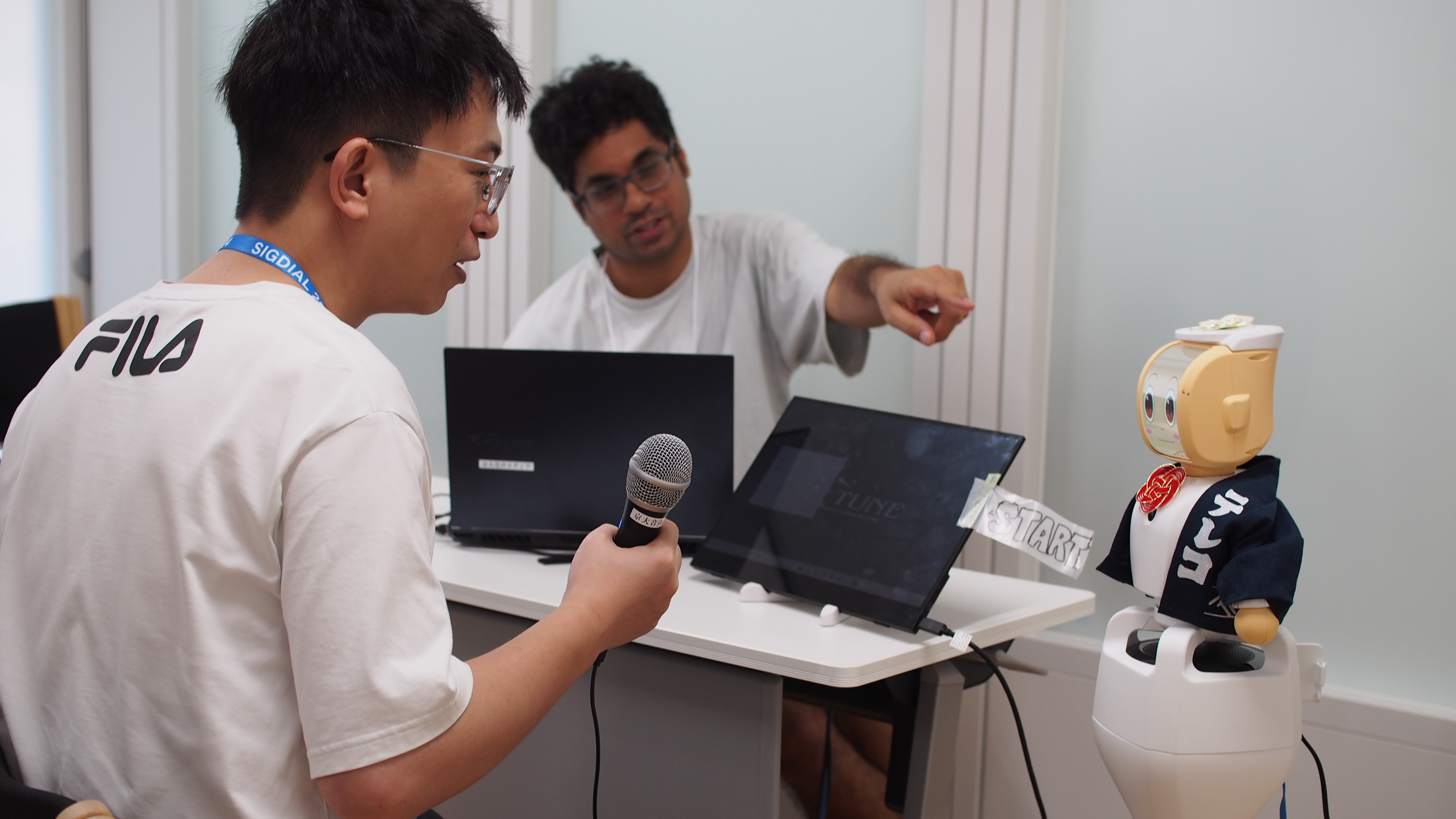} 
  \Description{Photo of interview dialogue with TELECO by SIGDIAL participant}
  \caption{Photo of interview dialogue with TELECO by SIGDIAL participant}
  \label{fig:teleco_photo}
\end{figure}

In this study, we employed two robots with different levels of human-likeness:
\begin{itemize}
\item \textbf{ERICA \cite{glas2016erica, inoue2016talking, kawahara2019spoken}:} An android robot designed to resemble an adult female (see Figure \ref{fig:erica_photo}).
\item \textbf{TELECO \cite{horikawa2023cybernetic, kondo2023practical}:} A humanoid robot featuring an OLED display for its face and simplified joint structures (see Figure \ref{fig:teleco_photo}). 
\end{itemize}


\begin{figure*}[h]
  \centering
  \includegraphics[width=\linewidth]{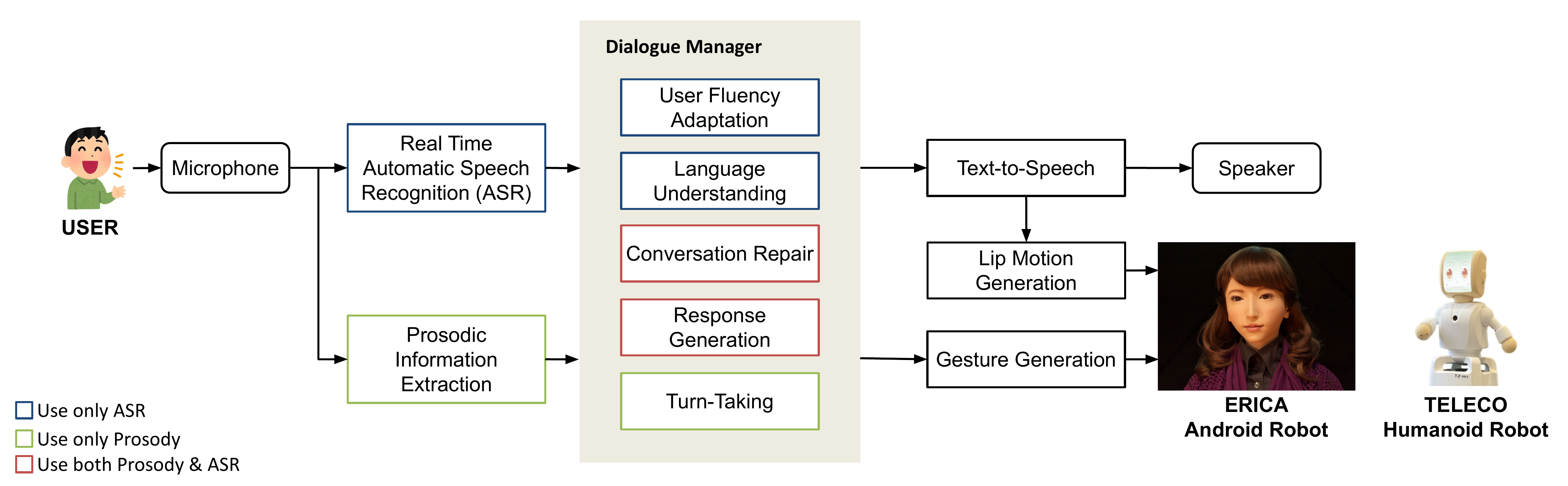} 
  \Description{Overall architecture of the interview system implemented in our study. This comprehensive system architecture includes modules for real-time automatic speech recognition (ASR), prosodic information extraction, language understanding, and user fluency adaptation, among others. Central to the system is the dialogue manager, which coordinates turn-taking, response generation, and conversation repair. Also included are the text-to-speech, gesture generation, and lip motion generation components, enhancing the robots' interactive capabilities.}
  \caption{Overall architecture of the interview system implemented in our study. This comprehensive system architecture includes modules for real-time automatic speech recognition (ASR), prosodic information extraction, language understanding, and user fluency adaptation, among others. Central to the system is the dialogue manager, which coordinates turn-taking, response generation, and conversation repair. Also included are the text-to-speech, gesture generation, and lip motion generation components, enhancing the robots' interactive capabilities.}
  \label{fig:architecture}
\end{figure*}

To isolate the impact of appearance on user behavior, we implemented the same human-like spoken dialogue system for both robots \cite{pang2024human}. They used identical dialogue behaviors, gestures, and facial expressions to ensure any observed differences could be attributed primarily to the differences in appearances. The system architecture can be found in Figure \ref{fig:architecture}. 

\subsection{Data Collection}
We conducted our study at SIGDIAL 2024\footnote{\url{https://2024.sigdial.org/}}, an international conference attended by over 160 participants. As shown in Figure \ref{fig:erica_photo} and Figure \ref{fig:teleco_photo}, attendees took part in brief, one-on-one interviews with either ERICA or TELECO, each lasting two to three minutes. We selected an interview approach because it is both engaging and well-suited to the conference setting, allowing us to observe how human-likeness might influence user behavior under natural interaction conditions.

No formal questionnaires were administered to maintain a casual atmosphere and natural interaction experience. Consent was obtained through informing the attendees about the study at the conference's opening session and through clearly displayed notices in the interview room, that only the transcripted dialogue, captured by Automatic Speech Recognition (ASR), would be recorded. Accordingly, our analysis focuses on user spoken behavior extracted from the interview transcripts to examine how variations in the robots' appearances affect user interactions.

\section{User Behavior Analysis} \label{sec:analysis}

\subsection{Behavior Metrics} \label{behavior} We investigated multiple dimensions of user spoken behavior, each reflecting different linguistic constructs. We grouped the metrics into four main categories—(1) Linguistic, (2) Dialogue, (3) Emotion, and (4) Behavioral Mimicry. Below, we outline the primary measures:

\paragraph{(1) Linguistic} This category quantifies the structural and lexical complexity of user responses: 
\begin{itemize}
    \item \textit{Number of Words and Utterance Length}: Total word count and the average number of words per utterance, indicating how extensively or succinctly participants responded. 
    \item \textit{Lexical Diversity}: The ratio of unique words to total words, widely considered a hallmark of expressive vocabulary and verbal fluency. 
    \item \textit{Syntactic Complexity}: Measured by dependency-parse depth through SpaCy\footnote{\url{https://github.com/explosion/spaCy}} , reflecting how participants layered or embedded phrases within each sentence (a higher average suggests more complex syntax). 
\end{itemize}

\paragraph{(2) Dialogue} This component examines user interactions and response dynamics:
\begin{itemize}
    \item \textit{Number of disfluencies and disfluencies ratio}: Counting interjections (e.g., ``um,'' ``uh'') and consecutive repeated words. Interjections is calculated using the \texttt{en\_core\_web\_lg} model\footnote{\url{https://spacy.io/models/en\#en_core_web_lg}} in spaCy \footnote{\url{https://github.com/explosion/spaCy}}. Repeated words is calculated through bi-gram models in NLTK library\footnote{\url{https://github.com/nltk/nltk}}.
    \item \textit{Politeness Score}: We evaluated whether user politeness was influenced by the robots appearance. The score is derived from an XLM-RoBERTa-based classification of polite vs. impolite utterances \cite{srinivasan2022tydip}, normalized between 0 and 1. 
    \item \textit{Word Commonness}: We evaluated how common word usage was influenced by the robots appearance. By using the Brown Corpus \cite{bird2009natural} as a reference corpus, we built a commonness score calculator based on the normalized frequency. Then we computed the final score through the average commonness score for all words in a single utterance, indicating the  ``commonness" of a user's vocabulary.
    \item \textit{Personal Pronouns Ratio}: We tracked whether users referred to the robot using personal pronouns (e.g., ``you,'' ``he,'' ``she''), or impersonal pronouns (e.g., ``it,'' ``the robot''), which can shed light on animacy attribution and self-reference patterns \cite{packard2018m, qu2022effect}. 
    \item \textit{Hedging Word Ratio}: We evaluated user hedging behavior based the robot's appearance. Hedging involves words or phrases expressing uncertainty or non-commitment (e.g., ``maybe,'' ``seem,'' ``usually''). We detected such terms using a predefined list\footnote{\url{https://github.com/words/hedges}}.
\end{itemize}

\paragraph{(3) Emotion} Focused on the emotional content of user interactions: 
\begin{itemize} 
    \item \textit{Sentiment Score}: To evaluate whether users' utterances conveyed positive or negative sentiment, we used SiEBERT \cite{hartmann2023}. We computed the sentiment score through polarity calculation (positive sentiment score - negative sentiment score). We then normalized the value to a range between 0 and 1, where higher scores indicate more positive sentiment. 
    \item \textit{Emotion Score (Joy, Sadness, Anger, Fear, Disgust, and Surprise)}: We implemented the likelihood estimates assigned by a DistilRoBERTa-based emotion classifier \cite{hartmann2022emotionenglish}, clarifying which emotions predominated in each user's speech. 
\end{itemize}

\paragraph{(4) Behavioral Mimicry} 
Investigates the extent of user mimicry of robot behavior, a key indicator of empathy and social bonding \cite{schulte2007mirror, gallese2006embodied, praszkier2016empathy}:
\begin{itemize}
\item \textit{Lexical Mirroring}: Overlap in word choice between the user's and robot's utterances, normalized by the user's total number of words. 
\item \textit{Semantic Mirroring}: A BERTScore-based measure indicating semantic similarity between user and robot dialogue. 
\item \textit{Syntactic Mirroring}: Assessed via POS-based cosine similarity, capturing how closely users' grammatical structures mirror the robot's output. 
\end{itemize}

\begin{table}[h]
\centering
{\small
\begin{tabular}{lrrr}
\hline
Behavior                                                          & ERICA          & TELECO         & Effect Size \\ \hline
\rowcolor[gray]{0.95}(Linguistic)            &                &                &             \\
\# of Words                                                       & 143.46 (66.17) & 109.39 (50.81) & 0.31        \\
Utterance Length                                                  & 10.32 (5.02)   & 8.48 (3.92)    & 0.17        \\
Lexical Diversity                                                 & 0.59 (0.09)    & 0.61 (0.09)    & 0.12        \\
Syntactic Complexity                                              & 2.81 (1.01)    & 2.20 (0.50)    & 0.41        \\
\rowcolor[gray]{0.95}(Dialogue)              &                &                &             \\
\# of Disfluencies                                                  & 7.92 (6.51)    & 12.78 (8.89)   & 0.42        \\
Disfluencies Ratio                                                  & 0.61 (0.59)    & 1.05 (0.78)    & 0.41        \\
Politeness Score                                                  & 0.63 (0.15)    & 0.60 (0.13)    & 0.10        \\
Word Commonness                                                   & 6.11 (0.40)    & 5.91 (0.44)    & 0.26        \\
Personal Pronouns Ratio                                           & 2.74 (3.23)    & 1.89 (1.64)    & 0.10        \\
Hedging Word Ratio                                                & 1.15 (0.63)    & 0.94 (0.59)    & 0.22        \\ 
\rowcolor[gray]{0.95}(Emotion)               &                &                &             \\
Sentiment Score                                                   & 0.67 (0.18)    & 0.64 (0.15)    & 0.13        \\
Joy                                                               & 0.14 (0.08)    & 0.11 (0.09)    & 0.28        \\
Sadness                                                           & 0.03 (0.02)    & 0.04 (0.03)    & 0.33        \\
Anger                                                             & 0.06 (0.02)    & 0.07 (0.02)    & 0.21        \\
Fear                                                              & 0.19 (0.08)    & 0.20 (0.07)    & 0.15        \\
Disgust                                                           & 0.08 (0.03)    & 0.08 (0.04)    & 0.05        \\
Surprise                                                          & 0.03 (0.02)    & 0.03 (0.02)    & 0.02        \\
\rowcolor[gray]{0.95}(Behavioral Mimicry) & \textbf{}      & \textbf{}      &             \\
Lexical Mirroring                                                 & 0.36 (0.07)    & 0.35 (0.11)    & 0.02        \\
Semantic Mirroring                                                & 0.51 (0.02)    & 0.50 (0.02)    & 0.19        \\
Syntactic Mirroring                                               & 0.29 (0.09)    & 0.28 (0.08)    & 0.01        \\

\hline
\end{tabular}
}
\caption{Comparative analysis of user behaviors based on robot appearance. Data presented includes the mean and standard deviation (in parentheses) for each behavior metric across the robots, accompanied by the computed effect sizes through Rank-biserial correlations.}
\label{tab:distribution}
\end{table}

\subsection{Result} \label{sec:analysis_result}
A total of 42 participants interacted with our robots during the conference, 24 with ERICA and 18 with TELECO. Results are detailed in Table \ref{tab:distribution}. Given our small sample size, we focused on practical significance, assessing effect sizes using Rank-biserial correlations rather than relying solely on statistical significance, which is less sensitive to sample size limitations \cite{sullivan2012using}. Additionally, we performed Mann-Whitney U tests; several features such as number of disfluencies (p = 0.022), disfluency ratio (p = 0.027), and syntactic complexity (p = 0.024) showed moderate unadjusted p-values. However, achieving statistical significance after Bonferroni correction ($\alpha$ = 0.002) was challenging due to multiple comparisons (21 in total).

Key findings include notable differences in disfluencies and syntactic complexity between interactions with ERICA and TELECO. Participants exhibited more disfluencies with TELECO both in total count (12.78 vs. 7.92, effect size = 0.417) and ratio (1.05 vs. 0.61, effect size = 0.405). They also used more complex syntax with ERICA (2.81 vs. 2.20, effect size = 0.405). Lesser variations were observed in the number of words (143.46 vs. 109.39, effect size = 0.31), word commonness (6.11 vs. 5.91, effect size = 0.26), and hedging word ratio (1.15 vs. 0.94, effect size = 0.22).

Regarding affective content, both groups reported positive sentiment scores (0.67 vs. 0.64), with slightly higher joy levels observed with ERICA (0.14 vs. 0.11). Emotional expressions such as anger, sadness, fear, surprise, and disgust showed minimal differences between the groups. Behavioral mimicry across lexical, semantic, and syntactic dimensions showed negligible differences (effect sizes = 0.02, 0.19, and 0.01, respectively).

\section{Predictive Model} \label{sec:model}

\subsection{Experimental Setup} 
In addition to user behavior analysis, we developed a predictive model to identify whether a participant interacted with the more human-like robot (ERICA) or the less human-like robot (TELECO), based solely on metrics of spoken behavior detailed in Section~\ref{behavior}. We selected input features for the model based on their differences and effect sizes both exceeding 0.1, which include the number of words, utterance length, syntactic complexity, number and ratio of disfluencies, word commonness, and hedging word ratio, highlighting discernible variations between the two datasets. Results using all user behavior metrics as input features are presented in Table \ref{tab:full_feature} in \ref{appendix:A} as part of an ablation study.

We evaluated a range of machine learning classifiers, including Random Forest \cite{breiman2001random}, Gradient Boosting\cite{friedman2002stochastic}, and Naïve Bayes \cite{rish2001empirical}, along with a random baseline, employing default hyperparameters for each algorithm. The dataset was divided into an 80\% training set and a 20\% test set. To reduce bias and validate model robustness, we conducted 3-fold cross-validation on the training set and repeated the training/validation process ten times with different random seeds to ensure reproducibility and assess stability. We averaged the performance metrics—accuracy, macro-average precision, recall, and F1-score—across folds and seeds, compiling the final performance metrics. Detailed results for each seed are documented in Table~\ref{tab:full_result} in Appendix~\ref{appendix:A}.

\begin{table}[h]
\begin{tabular}{lrrrr}
\hline
Model               & \multicolumn{1}{l}{Accuracy} & \multicolumn{1}{l}{Precision} & \multicolumn{1}{l}{Recall} & \multicolumn{1}{l}{F1-score} \\ \hline
Random Baseline     & 48.57                        & 48.29                         & 47.94                      & 46.76                        \\
Random Forest       & 64.29                        & 64.33                         & 64.20                      & 63.32                        \\
Gradient Boosting   & 63.10                        & 63.56                         & 63.21                      & 61.98                        \\
Naïve Bayes         & \textbf{72.38}               & \textbf{73.49}                & \textbf{72.79}             & \textbf{71.60}               \\ \hline
\end{tabular}
\caption{Predictive Model Evaluation Result [\%]}
\label{tab:result}
\end{table}

\subsection{Feature Importance Analysis} 
In addition to model development, we explored how specific features influence our predictive models using two suggested interpretability methods \cite{molnar2020general}:

\paragraph{Permutation Feature Importance (PFI) \cite{altmann2010permutation}} This model-agnostic technique evaluates the impact of individual features by measuring the reduction in model performance (e.g., F1-score) when the values of a feature are randomly shuffled in the test set. Significant performance drops indicate critical feature importance.

\paragraph{SHapley Additive exPlanations (SHAP) \cite{NIPS2017_7062}} SHAP provides a detailed measure of each feature's contribution to individual predictions, where we use the TreeExplainer \cite{lundberg2020local2global} to calculate SHAP values in this study.

\subsection{Results} \label{sec:model_result} Table~\ref{tab:result} shows that the Naïve Bayes model outperformed all other evaluated algorithms with the highest F1-score (71.60\%), significantly exceeding the random baseline (46.76\%) and other methods such as Random Forest, and Gradient Boosting (61-63\%). This underscores that linguistic features effectively differentiate participants' perceptions of robot human-likeness.

\begin{figure*}
  \centering
  \includegraphics[width=.75\linewidth]{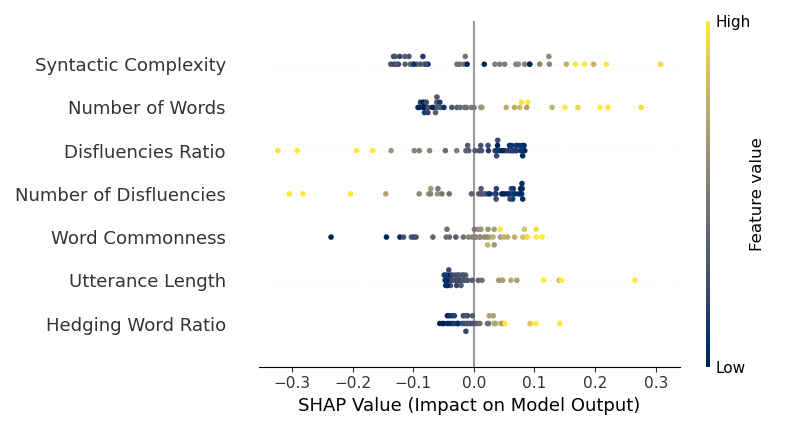} 
  \Description{Distribution of SHAP values for each behavioral feature. The SHAP values, which quantify the impact on the model output, are plotted along the x-axis against each feature on the y-axis. Each point represents an individual instance. Points in \textcolor{blue}{blue} indicate feature values below the average, affecting the model negatively (left of the vertical zero line), while points in \textcolor{yellow}{yellow} denote values above the average, contributing positively to the prediction (right of the zero line).}
  \caption{Distribution of SHAP values for each behavioral feature. The SHAP values, which quantify the impact on the model output, are plotted along the x-axis against each feature on the y-axis. Each point represents an individual instance. Points in \textcolor{blue}{blue} indicate feature values below the average, affecting the model negatively (left of the vertical zero line), while points in \textcolor{yellow}{yellow} denote values above the average, contributing positively to the prediction (right of the zero line).
}
  \label{fig:shap}
\end{figure*}

Feature impact analysis on the Naïve Bayes model revealed that syntactic complexity was the most influential predictor, contributing 22.51\% in SHAP and 0.079 in PFI. Other significant features included the number of words (17.17\% in SHAP, 0.047 in PFI) and various disfluency measures, with number and ratio contributing 15.34\% and 15.57\% in SHAP and 0.031 and 0.026 in PFI, respectively. These insights are visualized in Figures~\ref{fig:shap} and summarized in Table~\ref{tab:shap}.

\begin{table}[ht]
{\small
\begin{tabular}{lrrr}
\hline
\multicolumn{1}{c}{\multirow{2}{*}{Behavior}} & \multicolumn{1}{c}{\multirow{2}{*}{\begin{tabular}[c]{@{}c@{}}Permutation Feature \\ Importance (PFI)\end{tabular}}} & \multicolumn{2}{c}{SHAP}                                  \\ \cline{3-4} 
\multicolumn{1}{c}{}                          & \multicolumn{1}{c}{}                                                                                           & \multicolumn{1}{c}{Mean} & \multicolumn{1}{c}{Percentage} \\ \hline
Syntactic Complexity                          & 0.079 (0.050)                                                                                                  & 0.102                    & 22.51                          \\
Number of Words                               & 0.047 (0.035)                                                                                                  & 0.078                    & 17.17                          \\
Disfluencies Ratio                            & 0.026 (0.036)                                                                                                  & 0.070                    & 15.57                          \\
Number of Disfluencies                        & 0.031 (0.035)                                                                                                  & 0.069                    & 15.34                          \\
Word Commonness                               & 0.053 (0.041)                                                                                                  & 0.056                    & 12.32                          \\
Utterance Length                              & 0.003 (0.028)                                                                                                  & 0.045                    & 10.02                          \\
Hedging Word Ratio                            & 0.003 (0.031)                                                                                                  & 0.032                    & 7.07                           \\ \hline
\end{tabular}
}
\caption{Feature importance analysis for the Naïve Bayes model using Permutation Feature Importance (PFI) and SHapley Additive exPlanations (SHAP). The table displays the mean SHAP values and their respective contribution percentages, along with the PFI scores (mean and standard deviation in parentheses), sorted by their contribution to the model's predictive performance.}
\label{tab:shap}
\end{table}

\section{General Discussion}

This section reflects on key findings from the analysis and modeling results concerning disfluencies and syntactic complexity metrics. Notably, users engaged with ERICA, the more human-like robot, exhibited more complex syntax as detailed in Section \ref{sec:analysis_result}. This aligns with the predictive model results in Section \ref{sec:model_result}, where syntactic complexity emerged as a significant predictor. This observation may be attributed to the perceived cognitive capabilities of the robots, influenced by their level of human-likeness. According to Communication Accommodation Theory, individuals tend to adjust their communication style to match their conversational partner's perceived attributes or capabilities \cite{giles2007communication}. Previous studies also indicate that more human-like robots are often ascribed higher cognitive functions \cite{fortunati2023exploring}. Thus, ERICA with a human-like appearance will lead users to adopt more complex language, akin to human-human interactions, under the assumption that ERICA can process such communication.

In contrast, disfluencies demonstrated an opposite trend, with users exhibiting more disfluencies when interacting with the less human-like robot, TELECO. This phenomenon may be attributed to increased cognitive load, which refers to the utilization of working memory resources during cognitive tasks. Cognitive psychology suggests that unfamiliar tasks require more information processing, which can escalate cognitive load. Prior research has indicated that heightened cognitive load can manifest through various temporal characteristics such as altered speech rate and increased ratio of pauses, or through interactions with less legible text, leading to a rise in disfluency rates \cite{bona2020effect, kuhl2016effects}. In our context, TELECO's less human-like appearance might have been perceived as unfamiliar or unnatural compared to typical human interactions, necessitating greater cognitive effort and consequently, higher disfluency rates. This observation aligns with previous findings where interactions with autonomous systems, characterized by less human-like features, elicited more disfluencies than more anthropomorphic Wizard-of-Oz (WoZ) conditions \cite{elmers2024analysis}.

An intriguing outcome from our study relates to the behavior mimicry in Section \ref{sec:analysis_result}, where no significant differences in mimicry levels were found across human-likeness levels, contrasting with prior research that suggests humans often mimic robots during interactions \cite{suman2016prior, riek2010my}. This deviation could be explained by the context of the interactions, which were structured as interviews rather than dynamic social engagements. Previous literature indicates that mimicry is more prevalent in emotionally charged real-life interactions, where empathy and social bonding are more critical \cite{schulte2007mirror, gallese2006embodied, praszkier2016empathy, ravreby2024many}. The controlled interview setting of our study likely limited emotional engagement, thus reducing the occurrence of mimicry.

\section{Conclusion}

This study investigated the impact of a robot's human-likeness appearance on user speech patterns during real-world, fully autonomous interactions at an international conference. Data from 42 participants, engaging with the highly anthropomorphic ERICA and the more basic TELECO, were analyzed and modeled. The results align across experiments, showing that linguistic markers such as disfluencies, syntactic complexity, and utterance length vary noticeably between interactions with the two robots, demonstrating moderate effect sizes and substantial contributions in feature importance analyses. These findings suggest that these linguistic features are influential in shaping user perceptions of a robot's human-likeness. We further explore the relationship between these features and robot human-like appearance from a cognitive science perspective, linking increased syntactic complexity with more human-like robots via Communication Accommodation Theory, and higher disfluencies with less human-like robots due to greater cognitive load.

Looking forward, this research could be extended in several ways. Increasing the sample size and employing more controlled manipulations of robot appearances could further validate these initial findings. Moreover, exploring factors beyond appearance—such as enhanced AI capabilities or different interaction methods (Wizard-of-Oz vs. autonomous)—could enrich our understanding of human-likeness in robots. Additionally, expanding our analytical modalities to include non-verbal cues like facial expressions and speech tone could offer a more comprehensive view of how robot appearance affects human interactions. Integrating such non-verbal signals would provide a broader context for understanding the dynamic between robot embodiments and human behavior. Ultimately, the nuanced understanding of syntax and disfluencies revealed by our study could inform the design of robots that more effectively mirror human communicative norms, thereby improving both the perceived humanness and the quality of interactions.




\appendix

\section*{Appendix} \label{appendix:A}

\begin{table}[H]
\begin{tabular}{lrrrr}
\hline
Model               & \multicolumn{1}{l}{Accuracy} & \multicolumn{1}{l}{Precision} & \multicolumn{1}{l}{Recall} & \multicolumn{1}{l}{F1-score} \\ \hline
Random Baseline     & 49.76                        & 50.89                         & 50.33                      & 48.72                        \\
Random Forest       & 55.71                        & 55.25                         & 54.81                      & 53.27                        \\
Gradient Boosting   & 53.57                        & 52.59                         & 52.55                      & 50.73                        \\
Naive Bayes         & \textbf{57.62}                        & \textbf{59.35}                & \textbf{58.74}             & \textbf{56.59}               \\ \hline
\end{tabular}
\caption{Performance evaluation of predictive models [\%] without feature selection. Best performing values for each metric are highlighted in \textbf{bold}.}
\label{tab:full_feature}
\end{table}

\pagebreak

\begin{table}[H]
\begin{tabular}{lrrrr}
\hline
Model & Accuracy & Precision & Recall & F1-score \\
\hline
\rowcolor[gray]{0.95}Seed 1              &               &               &               &               \\
Random Baseline     & 0.33          & 0.34          & 0.33          & 0.33          \\
Random Forest       & 0.69          & 0.69          & 0.69          & 0.69          \\
Gradient Boosting   & 0.69          & 0.69          & 0.69          & 0.68          \\
Naive Bayes         & \textbf{0.76} & \textbf{0.79} & \textbf{0.78} & \textbf{0.76} \\
\rowcolor[gray]{0.95}Seed 2              &               &               &               &               \\
Random Baseline     & 0.52          & 0.57          & 0.54          & 0.51          \\
Random Forest       & 0.60          & 0.61          & 0.61          & 0.59          \\
Gradient Boosting   & 0.62          & 0.59          & 0.59          & 0.59          \\
Naive Bayes         & \textbf{0.74} & \textbf{0.74} & \textbf{0.73} & \textbf{0.73} \\
\rowcolor[gray]{0.95}Seed 3              &               &               &               &               \\
Random Baseline     & 0.55          & 0.57          & 0.56          & 0.53          \\
Random Forest       & 0.60          & 0.59          & 0.58          & 0.57          \\
Gradient Boosting   & 0.62          & 0.62          & 0.61          & 0.60          \\
Naive Bayes         & \textbf{0.74} & \textbf{0.81} & \textbf{0.75} & \textbf{0.72} \\
\rowcolor[gray]{0.95}Seed 4              &               &               &               &               \\
Random Baseline     & 0.48          & 0.44          & 0.45          & 0.44          \\
Random Forest       & \textbf{0.74} & \textbf{0.73} & \textbf{0.73} & \textbf{0.73} \\
Gradient Boosting   & 0.64          & 0.67          & 0.66          & 0.63          \\
Naive Bayes         & 0.69          & 0.70          & 0.70          & 0.69          \\ 
\rowcolor[gray]{0.95}Seed 5              &               &               &               &               \\
Random Baseline     & 0.48          & 0.46          & 0.47          & 0.46          \\
Random Forest       & 0.64          & 0.65          & 0.65          & 0.64          \\
Gradient Boosting   & 0.62          & 0.63          & 0.62          & 0.62          \\
Naive Bayes         & \textbf{0.71} & \textbf{0.72} & \textbf{0.72} & \textbf{0.71} \\
\rowcolor[gray]{0.95}Seed 6              &               &               &               &               \\
Random Baseline     & 0.40          & 0.40          & 0.41          & 0.40          \\
Random Forest       & 0.67          & 0.67          & 0.67          & 0.66          \\
Gradient Boosting   & 0.69          & 0.69          & 0.69          & 0.69          \\
Naive Bayes         & \textbf{0.71} & \textbf{0.72} & \textbf{0.72} & \textbf{0.71} \\
\rowcolor[gray]{0.95}Seed 7              &               &               &               &               \\
Random Baseline     & 0.40          & 0.38          & 0.38          & 0.38          \\
Random Forest       & 0.64          & 0.64          & 0.64          & 0.63          \\
Gradient Boosting   & 0.67          & 0.68          & 0.67          & 0.66          \\
Naive Bayes         & \textbf{0.71} & \textbf{0.72} & \textbf{0.73} & \textbf{0.71} \\
\rowcolor[gray]{0.95}Seed 8              &               &               &               &               \\
Random Baseline     & 0.45          & 0.46          & 0.46          & 0.42          \\
Random Forest       & 0.57          & 0.56          & 0.56          & 0.56          \\
Gradient Boosting   & 0.50          & 0.51          & 0.51          & 0.50          \\
Naive Bayes         & \textbf{0.71} & \textbf{0.72} & \textbf{0.72} & \textbf{0.71} \\
\rowcolor[gray]{0.95}Seed 9              &               &               &               &               \\
Random Baseline     & 0.69          & 0.68          & 0.69          & 0.68          \\
Random Forest       & \textbf{0.71} & \textbf{0.71} & \textbf{0.72} & \textbf{0.71} \\
Gradient Boosting   & \textbf{0.71} & \textbf{0.71} & 0.71          & 0.70          \\
Naive Bayes         & \textbf{0.71} & 0.70          & 0.70          & 0.70          \\
\rowcolor[gray]{0.95}Seed 10             &               &               &               & 
             \\
 Random Baseline     & 0.55          & 0.53          & 0.52          & 0.52          \\
Random Forest       & 0.57          & 0.57          & 0.57          & 0.56          \\
Gradient Boosting   & 0.55          & 0.56          & 0.55          & 0.53          \\
Naive Bayes         & \textbf{0.74} & \textbf{0.72} & \textbf{0.72} & \textbf{0.72} \\
\hline
\end{tabular}
\caption{Predictive model evaluation results [\%] for each random seed. The highest performing values for each metric per seed are \textbf{bolded}.}
\label{tab:full_result}
\end{table}


\begin{thebibliography}{66}


\ifx \showCODEN    \undefined \def \showCODEN     #1{\unskip}     \fi
\ifx \showDOI      \undefined \def \showDOI       #1{#1}\fi
\ifx \showISBNx    \undefined \def \showISBNx     #1{\unskip}     \fi
\ifx \showISBNxiii \undefined \def \showISBNxiii  #1{\unskip}     \fi
\ifx \showISSN     \undefined \def \showISSN      #1{\unskip}     \fi
\ifx \showLCCN     \undefined \def \showLCCN      #1{\unskip}     \fi
\ifx \shownote     \undefined \def \shownote      #1{#1}          \fi
\ifx \showarticletitle \undefined \def \showarticletitle #1{#1}   \fi
\ifx \showURL      \undefined \def \showURL       {\relax}        \fi
\providecommand\bibfield[2]{#2}
\providecommand\bibinfo[2]{#2}
\providecommand\natexlab[1]{#1}
\providecommand\showeprint[2][]{arXiv:#2}

\bibitem[Adiba et~al\mbox{.}(2021)]%
        {adiba2021towards}
\bibfield{author}{\bibinfo{person}{Amalia~Istiqlali Adiba}, \bibinfo{person}{Takeshi Homma}, {and} \bibinfo{person}{Toshinori Miyoshi}.} \bibinfo{year}{2021}\natexlab{}.
\newblock \showarticletitle{Towards immediate backchannel generation using attention-based early prediction model}. In \bibinfo{booktitle}{\emph{ICASSP 2021-2021 IEEE International Conference on Acoustics, Speech and Signal Processing (ICASSP)}}. IEEE, \bibinfo{pages}{7408--7412}.
\newblock


\bibitem[Ali et~al\mbox{.}(2020)]%
        {ali2020automatic}
\bibfield{author}{\bibinfo{person}{Ghazanfar Ali}, \bibinfo{person}{Myungho Lee}, {and} \bibinfo{person}{Jae-In Hwang}.} \bibinfo{year}{2020}\natexlab{}.
\newblock \showarticletitle{Automatic text-to-gesture rule generation for embodied conversational agents}.
\newblock \bibinfo{journal}{\emph{Computer Animation and Virtual Worlds}} \bibinfo{volume}{31}, \bibinfo{number}{4-5} (\bibinfo{year}{2020}), \bibinfo{pages}{e1944}.
\newblock


\bibitem[Altmann et~al\mbox{.}(2010)]%
        {altmann2010permutation}
\bibfield{author}{\bibinfo{person}{Andr{\'e} Altmann}, \bibinfo{person}{Laura Tolo{\c{s}}i}, \bibinfo{person}{Oliver Sander}, {and} \bibinfo{person}{Thomas Lengauer}.} \bibinfo{year}{2010}\natexlab{}.
\newblock \showarticletitle{Permutation importance: a corrected feature importance measure}.
\newblock \bibinfo{journal}{\emph{Bioinformatics}} \bibinfo{volume}{26}, \bibinfo{number}{10} (\bibinfo{year}{2010}), \bibinfo{pages}{1340--1347}.
\newblock


\bibitem[Bayramo{\u{g}}lu et~al\mbox{.}(2021)]%
        {bayramouglu2021engagement}
\bibfield{author}{\bibinfo{person}{{\"O}yk{\"u}~Zeynep Bayramo{\u{g}}lu}, \bibinfo{person}{Engin Erzin}, \bibinfo{person}{Tevfik~Metin Sezgin}, {and} \bibinfo{person}{Y{\"u}cel Yemez}.} \bibinfo{year}{2021}\natexlab{}.
\newblock \showarticletitle{Engagement rewarded actor-critic with conservative Q-learning for speech-driven laughter backchannel generation}. In \bibinfo{booktitle}{\emph{Proceedings of the 2021 International Conference on Multimodal Interaction}}. \bibinfo{pages}{613--618}.
\newblock


\bibitem[Belanche et~al\mbox{.}(2021)]%
        {belanche2021examining}
\bibfield{author}{\bibinfo{person}{Daniel Belanche}, \bibinfo{person}{Luis~V Casal{\'o}}, \bibinfo{person}{Jeroen Schepers}, {and} \bibinfo{person}{Carlos Flavi{\'a}n}.} \bibinfo{year}{2021}\natexlab{}.
\newblock \showarticletitle{Examining the effects of robots' physical appearance, warmth, and competence in frontline services: The Humanness-Value-Loyalty model}.
\newblock \bibinfo{journal}{\emph{Psychology \& Marketing}} \bibinfo{volume}{38}, \bibinfo{number}{12} (\bibinfo{year}{2021}), \bibinfo{pages}{2357--2376}.
\newblock


\bibitem[Bennett et~al\mbox{.}(2017)]%
        {bennett2017differences}
\bibfield{author}{\bibinfo{person}{Maxwell Bennett}, \bibinfo{person}{Tom Williams}, \bibinfo{person}{Daria Thames}, {and} \bibinfo{person}{Matthias Scheutz}.} \bibinfo{year}{2017}\natexlab{}.
\newblock \showarticletitle{Differences in interaction patterns and perception for teleoperated and autonomous humanoid robots}. In \bibinfo{booktitle}{\emph{2017 IEEE/RSJ International Conference on Intelligent Robots and Systems (IROS)}}. IEEE, \bibinfo{pages}{6589--6594}.
\newblock


\bibitem[Bird et~al\mbox{.}(2009)]%
        {bird2009natural}
\bibfield{author}{\bibinfo{person}{Steven Bird}, \bibinfo{person}{Ewan Klein}, {and} \bibinfo{person}{Edward Loper}.} \bibinfo{year}{2009}\natexlab{}.
\newblock \bibinfo{booktitle}{\emph{Natural language processing with Python: analyzing text with the natural language toolkit}}.
\newblock \bibinfo{publisher}{" O'Reilly Media, Inc."}.
\newblock


\bibitem[B{\'o}na and Bakti(2020)]%
        {bona2020effect}
\bibfield{author}{\bibinfo{person}{Judit B{\'o}na} {and} \bibinfo{person}{M{\'a}ria Bakti}.} \bibinfo{year}{2020}\natexlab{}.
\newblock \showarticletitle{The effect of cognitive load on temporal and disfluency patterns of speech: evidence from consecutive interpreting and sight translation}.
\newblock \bibinfo{journal}{\emph{Target}} \bibinfo{volume}{32}, \bibinfo{number}{3} (\bibinfo{year}{2020}), \bibinfo{pages}{482--506}.
\newblock


\bibitem[Breiman(2001)]%
        {breiman2001random}
\bibfield{author}{\bibinfo{person}{Leo Breiman}.} \bibinfo{year}{2001}\natexlab{}.
\newblock \showarticletitle{Random forests}.
\newblock \bibinfo{journal}{\emph{Machine learning}}  \bibinfo{volume}{45} (\bibinfo{year}{2001}), \bibinfo{pages}{5--32}.
\newblock


\bibitem[Chen et~al\mbox{.}(2021)]%
        {chen2021reasoning}
\bibfield{author}{\bibinfo{person}{Xiuying Chen}, \bibinfo{person}{Zhi Cui}, \bibinfo{person}{Jiayi Zhang}, \bibinfo{person}{Chen Wei}, \bibinfo{person}{Jianwei Cui}, \bibinfo{person}{Bin Wang}, \bibinfo{person}{Dongyan Zhao}, {and} \bibinfo{person}{Rui Yan}.} \bibinfo{year}{2021}\natexlab{}.
\newblock \showarticletitle{Reasoning in dialog: Improving response generation by context reading comprehension}. In \bibinfo{booktitle}{\emph{Proceedings of the AAAI Conference on Artificial Intelligence}}, Vol.~\bibinfo{volume}{35}. \bibinfo{pages}{12683--12691}.
\newblock


\bibitem[Elmers et~al\mbox{.}(2024)]%
        {elmers2024analysis}
\bibfield{author}{\bibinfo{person}{Mikey Elmers}, \bibinfo{person}{Koji Inoue}, \bibinfo{person}{Divesh Lala}, \bibinfo{person}{Keiko Ochi}, {and} \bibinfo{person}{Tatsuya Kawahara}.} \bibinfo{year}{2024}\natexlab{}.
\newblock \showarticletitle{Analysis and Detection of Differences in Spoken User Behaviors Between Autonomous and Wizard-of-Oz Systems}. In \bibinfo{booktitle}{\emph{2024 27th Conference of the Oriental COCOSDA International Committee for the Co-ordination and Standardisation of Speech Databases and Assessment Techniques (O-COCOSDA)}}. IEEE, \bibinfo{pages}{1--6}.
\newblock


\bibitem[Fortunati et~al\mbox{.}(2023)]%
        {fortunati2023exploring}
\bibfield{author}{\bibinfo{person}{Leopoldina Fortunati}, \bibinfo{person}{Anna~Maria Manganelli}, \bibinfo{person}{Joachim H{\"o}flich}, {and} \bibinfo{person}{Giovanni Ferrin}.} \bibinfo{year}{2023}\natexlab{}.
\newblock \showarticletitle{Exploring the perceptions of cognitive and affective capabilities of four, real, physical robots with a decreasing degree of morphological human likeness}.
\newblock \bibinfo{journal}{\emph{International Journal of Social Robotics}} \bibinfo{volume}{15}, \bibinfo{number}{3} (\bibinfo{year}{2023}), \bibinfo{pages}{547--561}.
\newblock


\bibitem[Friedman(2002)]%
        {friedman2002stochastic}
\bibfield{author}{\bibinfo{person}{Jerome~H Friedman}.} \bibinfo{year}{2002}\natexlab{}.
\newblock \showarticletitle{Stochastic gradient boosting}.
\newblock \bibinfo{journal}{\emph{Computational statistics \& data analysis}} \bibinfo{volume}{38}, \bibinfo{number}{4} (\bibinfo{year}{2002}), \bibinfo{pages}{367--378}.
\newblock


\bibitem[Fu et~al\mbox{.}(2023)]%
        {fu2023dual}
\bibfield{author}{\bibinfo{person}{Yahui Fu}, \bibinfo{person}{Koji Inoue}, \bibinfo{person}{Divesh Lala}, \bibinfo{person}{Kenta Yamamoto}, \bibinfo{person}{Chenhui Chu}, {and} \bibinfo{person}{Tatsuya Kawahara}.} \bibinfo{year}{2023}\natexlab{}.
\newblock \showarticletitle{Dual variational generative model and auxiliary retrieval for empathetic response generation by conversational robot}.
\newblock \bibinfo{journal}{\emph{Advanced Robotics}} \bibinfo{volume}{37}, \bibinfo{number}{21} (\bibinfo{year}{2023}), \bibinfo{pages}{1406--1418}.
\newblock


\bibitem[Gallese(2006)]%
        {gallese2006embodied}
\bibfield{author}{\bibinfo{person}{Vittorio Gallese}.} \bibinfo{year}{2006}\natexlab{}.
\newblock \showarticletitle{Embodied simulation: from mirror neuron systems to interpersonal relations}. In \bibinfo{booktitle}{\emph{Empathy and Fairness: Novartis Foundation Symposium 278}}. Wiley Online Library, \bibinfo{pages}{3--19}.
\newblock


\bibitem[Giles et~al\mbox{.}(2007)]%
        {giles2007communication}
\bibfield{author}{\bibinfo{person}{Howard Giles}, \bibinfo{person}{Tania Ogay}, {et~al\mbox{.}}} \bibinfo{year}{2007}\natexlab{}.
\newblock \showarticletitle{Communication accommodation theory}.
\newblock \bibinfo{journal}{\emph{Explaining communication: Contemporary theories and exemplars}} (\bibinfo{year}{2007}), \bibinfo{pages}{293--310}.
\newblock


\bibitem[Glas et~al\mbox{.}(2016)]%
        {glas2016erica}
\bibfield{author}{\bibinfo{person}{Dylan~F Glas}, \bibinfo{person}{Takashi Minato}, \bibinfo{person}{Carlos~T Ishi}, \bibinfo{person}{Tatsuya Kawahara}, {and} \bibinfo{person}{Hiroshi Ishiguro}.} \bibinfo{year}{2016}\natexlab{}.
\newblock \showarticletitle{Erica: The erato intelligent conversational android}. In \bibinfo{booktitle}{\emph{2016 25th IEEE International symposium on robot and human interactive communication (RO-MAN)}}. IEEE, \bibinfo{pages}{22--29}.
\newblock


\bibitem[Hartmann(2022)]%
        {hartmann2022emotionenglish}
\bibfield{author}{\bibinfo{person}{Jochen Hartmann}.} \bibinfo{year}{2022}\natexlab{}.
\newblock \bibinfo{title}{Emotion English DistilRoBERTa-base}.
\newblock \bibinfo{howpublished}{\url{https://huggingface.co/j-hartmann/emotion-english-distilroberta-base/}}.
\newblock


\bibitem[Hartmann et~al\mbox{.}(2023)]%
        {hartmann2023}
\bibfield{author}{\bibinfo{person}{Jochen Hartmann}, \bibinfo{person}{Mark Heitmann}, \bibinfo{person}{Christian Siebert}, {and} \bibinfo{person}{Christina Schamp}.} \bibinfo{year}{2023}\natexlab{}.
\newblock \showarticletitle{More than a Feeling: Accuracy and Application of Sentiment Analysis}.
\newblock \bibinfo{journal}{\emph{International Journal of Research in Marketing}} \bibinfo{volume}{40}, \bibinfo{number}{1} (\bibinfo{year}{2023}), \bibinfo{pages}{75--87}.
\newblock
\urldef\tempurl%
\url{https://doi.org/10.1016/j.ijresmar.2022.05.005}
\showDOI{\tempurl}


\bibitem[Horikawa et~al\mbox{.}(2023)]%
        {horikawa2023cybernetic}
\bibfield{author}{\bibinfo{person}{Yukiko Horikawa}, \bibinfo{person}{Takahiro Miyashita}, \bibinfo{person}{Akira Utsumi}, \bibinfo{person}{Shogo Nishimura}, {and} \bibinfo{person}{Satoshi Koizumi}.} \bibinfo{year}{2023}\natexlab{}.
\newblock \showarticletitle{Cybernetic avatar platform for supporting social activities of all people}. In \bibinfo{booktitle}{\emph{2023 IEEE/SICE International Symposium on System Integration (SII)}}. IEEE, \bibinfo{pages}{1--4}.
\newblock


\bibitem[Inoue et~al\mbox{.}(2024)]%
        {inoue2024yeah}
\bibfield{author}{\bibinfo{person}{Koji Inoue}, \bibinfo{person}{Divesh Lala}, \bibinfo{person}{Gabriel Skantze}, {and} \bibinfo{person}{Tatsuya Kawahara}.} \bibinfo{year}{2024}\natexlab{}.
\newblock \showarticletitle{Yeah, Un, Oh: Continuous and Real-time Backchannel Prediction with Fine-tuning of Voice Activity Projection}.
\newblock \bibinfo{journal}{\emph{arXiv preprint arXiv:2410.15929}} (\bibinfo{year}{2024}).
\newblock


\bibitem[Inoue et~al\mbox{.}(2016)]%
        {inoue2016talking}
\bibfield{author}{\bibinfo{person}{Koji Inoue}, \bibinfo{person}{Pierrick Milhorat}, \bibinfo{person}{Divesh Lala}, \bibinfo{person}{Tianyu Zhao}, {and} \bibinfo{person}{Tatsuya Kawahara}.} \bibinfo{year}{2016}\natexlab{}.
\newblock \showarticletitle{Talking with ERICA, an autonomous android}. In \bibinfo{booktitle}{\emph{Proceedings of the 17th Annual Meeting of the Special Interest Group on Discourse and Dialogue}}. \bibinfo{pages}{212--215}.
\newblock


\bibitem[Jung and Hahn(2023)]%
        {jung2023social}
\bibfield{author}{\bibinfo{person}{Yoonwon Jung} {and} \bibinfo{person}{Sowon Hahn}.} \bibinfo{year}{2023}\natexlab{}.
\newblock \showarticletitle{Social Robots As Companions for Lonely Hearts: The Role of Anthropomorphism and Robot Appearance}. In \bibinfo{booktitle}{\emph{2023 32nd IEEE International Conference on Robot and Human Interactive Communication (RO-MAN)}}. IEEE, \bibinfo{pages}{2520--2525}.
\newblock


\bibitem[Kanda et~al\mbox{.}(2008)]%
        {kanda2008analysis}
\bibfield{author}{\bibinfo{person}{Takayuki Kanda}, \bibinfo{person}{Takahiro Miyashita}, \bibinfo{person}{Taku Osada}, \bibinfo{person}{Yuji Haikawa}, {and} \bibinfo{person}{Hiroshi Ishiguro}.} \bibinfo{year}{2008}\natexlab{}.
\newblock \showarticletitle{Analysis of humanoid appearances in human--robot interaction}.
\newblock \bibinfo{journal}{\emph{IEEE transactions on robotics}} \bibinfo{volume}{24}, \bibinfo{number}{3} (\bibinfo{year}{2008}), \bibinfo{pages}{725--735}.
\newblock


\bibitem[Kawahara(2019)]%
        {kawahara2019spoken}
\bibfield{author}{\bibinfo{person}{Tatsuya Kawahara}.} \bibinfo{year}{2019}\natexlab{}.
\newblock \showarticletitle{Spoken dialogue system for a human-like conversational robot ERICA}. In \bibinfo{booktitle}{\emph{9th International Workshop on Spoken Dialogue System Technology}}. Springer, \bibinfo{pages}{65--75}.
\newblock


\bibitem[Kim et~al\mbox{.}(2019)]%
        {kim2019eliza}
\bibfield{author}{\bibinfo{person}{Seo~Young Kim}, \bibinfo{person}{Bernd~H Schmitt}, {and} \bibinfo{person}{Nadia~M Thalmann}.} \bibinfo{year}{2019}\natexlab{}.
\newblock \showarticletitle{Eliza in the uncanny valley: Anthropomorphizing consumer robots increases their perceived warmth but decreases liking}.
\newblock \bibinfo{journal}{\emph{Marketing letters}}  \bibinfo{volume}{30} (\bibinfo{year}{2019}), \bibinfo{pages}{1--12}.
\newblock


\bibitem[Kondo et~al\mbox{.}(2023)]%
        {kondo2023practical}
\bibfield{author}{\bibinfo{person}{Yoshihisa Kondo}, \bibinfo{person}{Hiroyuki Yomo}, \bibinfo{person}{Shogo Nishimura}, \bibinfo{person}{Akira Utsumi}, {and} \bibinfo{person}{Takahiro Miyashita}.} \bibinfo{year}{2023}\natexlab{}.
\newblock \showarticletitle{A Practical Implementation of Multi-Radio Wi-Fi for Teleoperated Mobile Robots}. In \bibinfo{booktitle}{\emph{2023 IEEE International Conference on Omni-layer Intelligent Systems (COINS)}}. IEEE, \bibinfo{pages}{1--6}.
\newblock


\bibitem[K{\"u}hl and Eitel(2016)]%
        {kuhl2016effects}
\bibfield{author}{\bibinfo{person}{Tim K{\"u}hl} {and} \bibinfo{person}{Alexander Eitel}.} \bibinfo{year}{2016}\natexlab{}.
\newblock \showarticletitle{Effects of disfluency on cognitive and metacognitive processes and outcomes}.
\newblock \bibinfo{journal}{\emph{Metacognition and Learning}}  \bibinfo{volume}{11} (\bibinfo{year}{2016}), \bibinfo{pages}{1--13}.
\newblock


\bibitem[Li et~al\mbox{.}(2010)]%
        {li2010cross}
\bibfield{author}{\bibinfo{person}{Dingjun Li}, \bibinfo{person}{PL~Patrick Rau}, {and} \bibinfo{person}{Ye Li}.} \bibinfo{year}{2010}\natexlab{}.
\newblock \showarticletitle{A cross-cultural study: Effect of robot appearance and task}.
\newblock \bibinfo{journal}{\emph{International Journal of Social Robotics}}  \bibinfo{volume}{2} (\bibinfo{year}{2010}), \bibinfo{pages}{175--186}.
\newblock


\bibitem[Liang et~al\mbox{.}(2021)]%
        {liang2021infusing}
\bibfield{author}{\bibinfo{person}{Yunlong Liang}, \bibinfo{person}{Fandong Meng}, \bibinfo{person}{Ying Zhang}, \bibinfo{person}{Yufeng Chen}, \bibinfo{person}{Jinan Xu}, {and} \bibinfo{person}{Jie Zhou}.} \bibinfo{year}{2021}\natexlab{}.
\newblock \showarticletitle{Infusing multi-source knowledge with heterogeneous graph neural network for emotional conversation generation}. In \bibinfo{booktitle}{\emph{Proceedings of the AAAI Conference on Artificial Intelligence}}, Vol.~\bibinfo{volume}{35}. \bibinfo{pages}{13343--13352}.
\newblock


\bibitem[Likert(1932)]%
        {likert1932technique}
\bibfield{author}{\bibinfo{person}{Rensis Likert}.} \bibinfo{year}{1932}\natexlab{}.
\newblock \showarticletitle{A technique for the measurement of attitudes}.
\newblock \bibinfo{journal}{\emph{Archives of Psychology}} (\bibinfo{year}{1932}).
\newblock


\bibitem[Liu et~al\mbox{.}(2022)]%
        {liu2022friendly}
\bibfield{author}{\bibinfo{person}{Xing~Stella Liu}, \bibinfo{person}{Xiao~Shannon Yi}, {and} \bibinfo{person}{Lisa~C Wan}.} \bibinfo{year}{2022}\natexlab{}.
\newblock \showarticletitle{Friendly or competent? The effects of perception of robot appearance and service context on usage intention}.
\newblock \bibinfo{journal}{\emph{Annals of Tourism Research}}  \bibinfo{volume}{92} (\bibinfo{year}{2022}), \bibinfo{pages}{103324}.
\newblock


\bibitem[Lundberg et~al\mbox{.}(2020)]%
        {lundberg2020local2global}
\bibfield{author}{\bibinfo{person}{Scott~M. Lundberg}, \bibinfo{person}{Gabriel Erion}, \bibinfo{person}{Hugh Chen}, \bibinfo{person}{Alex DeGrave}, \bibinfo{person}{Jordan~M. Prutkin}, \bibinfo{person}{Bala Nair}, \bibinfo{person}{Ronit Katz}, \bibinfo{person}{Jonathan Himmelfarb}, \bibinfo{person}{Nisha Bansal}, {and} \bibinfo{person}{Su-In Lee}.} \bibinfo{year}{2020}\natexlab{}.
\newblock \showarticletitle{From local explanations to global understanding with explainable AI for trees}.
\newblock \bibinfo{journal}{\emph{Nature Machine Intelligence}} \bibinfo{volume}{2}, \bibinfo{number}{1} (\bibinfo{year}{2020}), \bibinfo{pages}{2522--5839}.
\newblock


\bibitem[Lundberg and Lee(2017)]%
        {NIPS2017_7062}
\bibfield{author}{\bibinfo{person}{Scott~M Lundberg} {and} \bibinfo{person}{Su-In Lee}.} \bibinfo{year}{2017}\natexlab{}.
\newblock \showarticletitle{A Unified Approach to Interpreting Model Predictions}.
\newblock In \bibinfo{booktitle}{\emph{Advances in Neural Information Processing Systems 30}}, \bibfield{editor}{\bibinfo{person}{I.~Guyon}, \bibinfo{person}{U.~V. Luxburg}, \bibinfo{person}{S.~Bengio}, \bibinfo{person}{H.~Wallach}, \bibinfo{person}{R.~Fergus}, \bibinfo{person}{S.~Vishwanathan}, {and} \bibinfo{person}{R.~Garnett}} (Eds.). \bibinfo{publisher}{Curran Associates, Inc.}, \bibinfo{pages}{4765--4774}.
\newblock
\urldef\tempurl%
\url{http://papers.nips.cc/paper/7062-a-unified-approach-to-interpreting-model-predictions.pdf}
\showURL{%
\tempurl}


\bibitem[Molnar et~al\mbox{.}(2020)]%
        {molnar2020general}
\bibfield{author}{\bibinfo{person}{Christoph Molnar}, \bibinfo{person}{Gunnar K{\"o}nig}, \bibinfo{person}{Julia Herbinger}, \bibinfo{person}{Timo Freiesleben}, \bibinfo{person}{Susanne Dandl}, \bibinfo{person}{Christian~A Scholbeck}, \bibinfo{person}{Giuseppe Casalicchio}, \bibinfo{person}{Moritz Grosse-Wentrup}, {and} \bibinfo{person}{Bernd Bischl}.} \bibinfo{year}{2020}\natexlab{}.
\newblock \showarticletitle{General pitfalls of model-agnostic interpretation methods for machine learning models}. In \bibinfo{booktitle}{\emph{International Workshop on Extending Explainable AI Beyond Deep Models and Classifiers}}. Springer, \bibinfo{pages}{39--68}.
\newblock


\bibitem[Nagy et~al\mbox{.}(2021)]%
        {nagy2021framework}
\bibfield{author}{\bibinfo{person}{Rajmund Nagy}, \bibinfo{person}{Taras Kucherenko}, \bibinfo{person}{Birger Moell}, \bibinfo{person}{Andr{\'e} Pereira}, \bibinfo{person}{Hedvig Kjellstr{\"o}m}, {and} \bibinfo{person}{Ulysses Bernardet}.} \bibinfo{year}{2021}\natexlab{}.
\newblock \showarticletitle{A framework for integrating gesture generation models into interactive conversational agents}.
\newblock \bibinfo{journal}{\emph{arXiv preprint arXiv:2102.12302}} (\bibinfo{year}{2021}).
\newblock


\bibitem[Niculescu et~al\mbox{.}(2013)]%
        {niculescu2013making}
\bibfield{author}{\bibinfo{person}{Andreea Niculescu}, \bibinfo{person}{Betsy Van~Dijk}, \bibinfo{person}{Anton Nijholt}, \bibinfo{person}{Haizhou Li}, {and} \bibinfo{person}{Swee~Lan See}.} \bibinfo{year}{2013}\natexlab{}.
\newblock \showarticletitle{Making social robots more attractive: the effects of voice pitch, humor and empathy}.
\newblock \bibinfo{journal}{\emph{International journal of social robotics}}  \bibinfo{volume}{5} (\bibinfo{year}{2013}), \bibinfo{pages}{171--191}.
\newblock


\bibitem[Packard et~al\mbox{.}(2018)]%
        {packard2018m}
\bibfield{author}{\bibinfo{person}{Grant Packard}, \bibinfo{person}{Sarah~G Moore}, {and} \bibinfo{person}{Brent McFerran}.} \bibinfo{year}{2018}\natexlab{}.
\newblock \showarticletitle{(I'm) happy to help (you): The impact of personal pronoun use in customer--firm interactions}.
\newblock \bibinfo{journal}{\emph{Journal of Marketing Research}} \bibinfo{volume}{55}, \bibinfo{number}{4} (\bibinfo{year}{2018}), \bibinfo{pages}{541--555}.
\newblock


\bibitem[Pang et~al\mbox{.}(2024a)]%
        {pang2024human}
\bibfield{author}{\bibinfo{person}{Zi~Haur Pang}, \bibinfo{person}{Yahui Fu}, \bibinfo{person}{Divesh Lala}, \bibinfo{person}{Mikey Elmers}, \bibinfo{person}{Koji Inoue}, {and} \bibinfo{person}{Tatsuya Kawahara}.} \bibinfo{year}{2024}\natexlab{a}.
\newblock \showarticletitle{Human-Like Embodied AI Interviewer: Employing Android ERICA in Real International Conference}.
\newblock \bibinfo{journal}{\emph{arXiv preprint arXiv:2412.09867}} (\bibinfo{year}{2024}).
\newblock


\bibitem[Pang et~al\mbox{.}(2024b)]%
        {pang2024acknowledgment}
\bibfield{author}{\bibinfo{person}{Zi~Haur Pang}, \bibinfo{person}{Yahui Fu}, \bibinfo{person}{Divesh Lala}, \bibinfo{person}{Keiko Ochi}, \bibinfo{person}{Koji Inoue}, {and} \bibinfo{person}{Tatsuya Kawahara}.} \bibinfo{year}{2024}\natexlab{b}.
\newblock \showarticletitle{Acknowledgment of Emotional States: Generating Validating Responses for Empathetic Dialogue}.
\newblock \bibinfo{journal}{\emph{arXiv preprint arXiv:2402.12770}} (\bibinfo{year}{2024}).
\newblock


\bibitem[Prakash and Rogers(2015)]%
        {prakash2015some}
\bibfield{author}{\bibinfo{person}{Akanksha Prakash} {and} \bibinfo{person}{Wendy~A Rogers}.} \bibinfo{year}{2015}\natexlab{}.
\newblock \showarticletitle{Why some humanoid faces are perceived more positively than others: effects of human-likeness and task}.
\newblock \bibinfo{journal}{\emph{International journal of social robotics}} \bibinfo{volume}{7}, \bibinfo{number}{2} (\bibinfo{year}{2015}), \bibinfo{pages}{309--331}.
\newblock


\bibitem[Praszkier(2016)]%
        {praszkier2016empathy}
\bibfield{author}{\bibinfo{person}{Ryszard Praszkier}.} \bibinfo{year}{2016}\natexlab{}.
\newblock \showarticletitle{Empathy, mirror neurons and SYNC}.
\newblock \bibinfo{journal}{\emph{Mind \& Society}}  \bibinfo{volume}{15} (\bibinfo{year}{2016}), \bibinfo{pages}{1--25}.
\newblock


\bibitem[Qu et~al\mbox{.}(2022)]%
        {qu2022effect}
\bibfield{author}{\bibinfo{person}{Jianhong Qu}, \bibinfo{person}{Ronggang Zhou}, {and} \bibinfo{person}{Zhe Chen}.} \bibinfo{year}{2022}\natexlab{}.
\newblock \showarticletitle{The effect of personal pronouns on users and the social role of conversational agents}.
\newblock \bibinfo{journal}{\emph{Behaviour \& Information Technology}} \bibinfo{volume}{41}, \bibinfo{number}{16} (\bibinfo{year}{2022}), \bibinfo{pages}{3470--3486}.
\newblock


\bibitem[Ravreby et~al\mbox{.}(2024)]%
        {ravreby2024many}
\bibfield{author}{\bibinfo{person}{Inbal Ravreby}, \bibinfo{person}{Mayan Navon}, \bibinfo{person}{Eliya Pinhas}, \bibinfo{person}{Jenya Lerer}, \bibinfo{person}{Yoav Bar-Anan}, {and} \bibinfo{person}{Yaara Yeshurun}.} \bibinfo{year}{2024}\natexlab{}.
\newblock \showarticletitle{The many faces of mimicry depend on the social context.}
\newblock \bibinfo{journal}{\emph{Emotion}} (\bibinfo{year}{2024}).
\newblock


\bibitem[Riek et~al\mbox{.}(2010)]%
        {riek2010my}
\bibfield{author}{\bibinfo{person}{Laurel~D Riek}, \bibinfo{person}{Philip~C Paul}, {and} \bibinfo{person}{Peter Robinson}.} \bibinfo{year}{2010}\natexlab{}.
\newblock \showarticletitle{When my robot smiles at me: Enabling human-robot rapport via real-time head gesture mimicry}.
\newblock \bibinfo{journal}{\emph{Journal on Multimodal User Interfaces}}  \bibinfo{volume}{3} (\bibinfo{year}{2010}), \bibinfo{pages}{99--108}.
\newblock


\bibitem[Rish et~al\mbox{.}(2001)]%
        {rish2001empirical}
\bibfield{author}{\bibinfo{person}{Irina Rish} {et~al\mbox{.}}} \bibinfo{year}{2001}\natexlab{}.
\newblock \showarticletitle{An empirical study of the naive Bayes classifier}. In \bibinfo{booktitle}{\emph{IJCAI 2001 workshop on empirical methods in artificial intelligence}}, Vol.~\bibinfo{volume}{3}. Seattle, WA, USA;, \bibinfo{pages}{41--46}.
\newblock


\bibitem[R{\i}zvano{\u{g}}lu et~al\mbox{.}(2014)]%
        {rizvanouglu2014impact}
\bibfield{author}{\bibinfo{person}{Kerem R{\i}zvano{\u{g}}lu}, \bibinfo{person}{{\"O}zg{\"u}rol {\"O}zt{\"u}rk}, {and} \bibinfo{person}{{\"O}ner Ad{\i}yaman}.} \bibinfo{year}{2014}\natexlab{}.
\newblock \showarticletitle{The impact of human likeness on the older adults’ perceptions and preferences of humanoid robot appearance}. In \bibinfo{booktitle}{\emph{Design, User Experience, and Usability. User Experience Design Practice: Third International Conference, DUXU 2014, Held as Part of HCI International 2014, Heraklion, Crete, Greece, June 22-27, 2014, Proceedings, Part IV 3}}. Springer, \bibinfo{pages}{164--172}.
\newblock


\bibitem[Saeki and Ueda(2024)]%
        {saeki2024impact}
\bibfield{author}{\bibinfo{person}{Waka Saeki} {and} \bibinfo{person}{Yoshiyuki Ueda}.} \bibinfo{year}{2024}\natexlab{}.
\newblock \showarticletitle{Impact of politeness and performance quality of android robots on future interaction decisions: a conversational design perspective}.
\newblock \bibinfo{journal}{\emph{Frontiers in Robotics and AI}}  \bibinfo{volume}{11} (\bibinfo{year}{2024}), \bibinfo{pages}{1393456}.
\newblock


\bibitem[Sasser et~al\mbox{.}(2024)]%
        {sasser2024investigation}
\bibfield{author}{\bibinfo{person}{Jordan~A Sasser}, \bibinfo{person}{Daniel~S McConnell}, {and} \bibinfo{person}{Janan~A Smither}.} \bibinfo{year}{2024}\natexlab{}.
\newblock \showarticletitle{Investigation of Relationships Between Embodiment Perceptions and Perceived Social Presence in Human--Robot Interactions}.
\newblock \bibinfo{journal}{\emph{International Journal of Social Robotics}} (\bibinfo{year}{2024}), \bibinfo{pages}{1--16}.
\newblock


\bibitem[Schulte-R{\"u}ther et~al\mbox{.}(2007)]%
        {schulte2007mirror}
\bibfield{author}{\bibinfo{person}{Martin Schulte-R{\"u}ther}, \bibinfo{person}{Hans~J Markowitsch}, \bibinfo{person}{Gereon~R Fink}, {and} \bibinfo{person}{Martina Piefke}.} \bibinfo{year}{2007}\natexlab{}.
\newblock \showarticletitle{Mirror neuron and theory of mind mechanisms involved in face-to-face interactions: a functional magnetic resonance imaging approach to empathy}.
\newblock \bibinfo{journal}{\emph{Journal of cognitive neuroscience}} \bibinfo{volume}{19}, \bibinfo{number}{8} (\bibinfo{year}{2007}), \bibinfo{pages}{1354--1372}.
\newblock


\bibitem[Song et~al\mbox{.}(2022)]%
        {song2022costume}
\bibfield{author}{\bibinfo{person}{Sichao Song}, \bibinfo{person}{Jun Baba}, \bibinfo{person}{Junya Nakanishi}, \bibinfo{person}{Yuichiro Yoshikawa}, {and} \bibinfo{person}{Hiroshi Ishiguro}.} \bibinfo{year}{2022}\natexlab{}.
\newblock \showarticletitle{Costume vs. Wizard of Oz vs. Telepresence: how social presence forms of tele-operated robots influence customer behavior}. In \bibinfo{booktitle}{\emph{2022 17th ACM/IEEE International Conference on Human-Robot Interaction (HRI)}}. IEEE, \bibinfo{pages}{521--529}.
\newblock


\bibitem[Srinivasan and Choi(2022)]%
        {srinivasan2022tydip}
\bibfield{author}{\bibinfo{person}{Anirudh Srinivasan} {and} \bibinfo{person}{Eunsol Choi}.} \bibinfo{year}{2022}\natexlab{}.
\newblock \showarticletitle{TyDiP: A Dataset for Politeness Classification in Nine Typologically Diverse Languages}. In \bibinfo{booktitle}{\emph{Findings of the Association for Computational Linguistics: EMNLP 2022}}. \bibinfo{pages}{5723--5738}.
\newblock


\bibitem[Straub(2016)]%
        {straub2016looks}
\bibfield{author}{\bibinfo{person}{Ilona Straub}.} \bibinfo{year}{2016}\natexlab{}.
\newblock \showarticletitle{‘It looks like a human!’The interrelation of social presence, interaction and agency ascription: a case study about the effects of an android robot on social agency ascription}.
\newblock \bibinfo{journal}{\emph{AI \& society}}  \bibinfo{volume}{31} (\bibinfo{year}{2016}), \bibinfo{pages}{553--571}.
\newblock


\bibitem[Sullivan and Feinn(2012)]%
        {sullivan2012using}
\bibfield{author}{\bibinfo{person}{Gail~M Sullivan} {and} \bibinfo{person}{Richard Feinn}.} \bibinfo{year}{2012}\natexlab{}.
\newblock \showarticletitle{Using effect size—or why the P value is not enough}.
\newblock \bibinfo{journal}{\emph{Journal of graduate medical education}} \bibinfo{volume}{4}, \bibinfo{number}{3} (\bibinfo{year}{2012}), \bibinfo{pages}{279--282}.
\newblock


\bibitem[Suman et~al\mbox{.}(2016)]%
        {suman2016prior}
\bibfield{author}{\bibinfo{person}{Apurv Suman}, \bibinfo{person}{Rebecca Marvin}, \bibinfo{person}{Elena~Corina Grigore}, \bibinfo{person}{Henny Admoni}, {and} \bibinfo{person}{Brian Scassellati}.} \bibinfo{year}{2016}\natexlab{}.
\newblock \showarticletitle{Prior behavior impacts human mimicry of robots}. In \bibinfo{booktitle}{\emph{2016 25th IEEE International Symposium on Robot and Human Interactive Communication (RO-MAN)}}. IEEE, \bibinfo{pages}{1057--1062}.
\newblock


\bibitem[Trovato et~al\mbox{.}(2016)]%
        {trovato2016hugging}
\bibfield{author}{\bibinfo{person}{Gabriele Trovato}, \bibinfo{person}{Martin Do}, \bibinfo{person}{{\"O}mer Terlemez}, \bibinfo{person}{Christian Mandery}, \bibinfo{person}{Hiroyuki Ishii}, \bibinfo{person}{Nadia Bianchi-Berthouze}, \bibinfo{person}{Tamim Asfour}, {and} \bibinfo{person}{Atsuo Takanishi}.} \bibinfo{year}{2016}\natexlab{}.
\newblock \showarticletitle{Is hugging a robot weird? Investigating the influence of robot appearance on users' perception of hugging}. In \bibinfo{booktitle}{\emph{2016 IEEE-RAS 16th International Conference on Humanoid Robots (Humanoids)}}. IEEE, \bibinfo{pages}{318--323}.
\newblock


\bibitem[Tung(2016)]%
        {tung2016child}
\bibfield{author}{\bibinfo{person}{Fang-Wu Tung}.} \bibinfo{year}{2016}\natexlab{}.
\newblock \showarticletitle{Child perception of humanoid robot appearance and behavior}.
\newblock \bibinfo{journal}{\emph{International Journal of Human-Computer Interaction}} \bibinfo{volume}{32}, \bibinfo{number}{6} (\bibinfo{year}{2016}), \bibinfo{pages}{493--502}.
\newblock


\bibitem[Villani et~al\mbox{.}(2024)]%
        {villani2024does}
\bibfield{author}{\bibinfo{person}{Alberto Villani}, \bibinfo{person}{T~Lisini Baldi}, \bibinfo{person}{Nicole D’Aurizio}, \bibinfo{person}{Giulio Campagna}, {and} \bibinfo{person}{Domenico Prattichizzo}.} \bibinfo{year}{2024}\natexlab{}.
\newblock \showarticletitle{Does Robot Anthropomorphism Improve Performance and User Experience in Teleoperation?}. In \bibinfo{booktitle}{\emph{2024 IEEE-RAS 23rd International Conference on Humanoid Robots (Humanoids)}}. IEEE, \bibinfo{pages}{76--83}.
\newblock


\bibitem[Walters et~al\mbox{.}(2008)]%
        {walters2008avoiding}
\bibfield{author}{\bibinfo{person}{Michael~L Walters}, \bibinfo{person}{Dag~S Syrdal}, \bibinfo{person}{Kerstin Dautenhahn}, \bibinfo{person}{Ren{\'e} Te~Boekhorst}, {and} \bibinfo{person}{Kheng~Lee Koay}.} \bibinfo{year}{2008}\natexlab{}.
\newblock \showarticletitle{Avoiding the uncanny valley: robot appearance, personality and consistency of behavior in an attention-seeking home scenario for a robot companion}.
\newblock \bibinfo{journal}{\emph{Autonomous Robots}}  \bibinfo{volume}{24} (\bibinfo{year}{2008}), \bibinfo{pages}{159--178}.
\newblock


\bibitem[Wang et~al\mbox{.}(2021)]%
        {wang2021empathetic}
\bibfield{author}{\bibinfo{person}{Jiashuo Wang}, \bibinfo{person}{Wenjie Li}, \bibinfo{person}{Peiqin Lin}, {and} \bibinfo{person}{Feiteng Mu}.} \bibinfo{year}{2021}\natexlab{}.
\newblock \showarticletitle{Empathetic response generation through graph-based multi-hop reasoning on emotional causality}.
\newblock \bibinfo{journal}{\emph{Knowledge-Based Systems}}  \bibinfo{volume}{233} (\bibinfo{year}{2021}), \bibinfo{pages}{107547}.
\newblock


\bibitem[Wu et~al\mbox{.}(2021)]%
        {wu2021personalized}
\bibfield{author}{\bibinfo{person}{Yuwei Wu}, \bibinfo{person}{Xuezhe Ma}, {and} \bibinfo{person}{Diyi Yang}.} \bibinfo{year}{2021}\natexlab{}.
\newblock \showarticletitle{Personalized response generation via generative split memory network}. In \bibinfo{booktitle}{\emph{Proceedings of the 2021 Conference of the North American Chapter of the Association for Computational Linguistics: Human Language Technologies}}. \bibinfo{pages}{1956--1970}.
\newblock


\bibitem[Yang and Xie(2024)]%
        {yang2024can}
\bibfield{author}{\bibinfo{person}{Wenjing Yang} {and} \bibinfo{person}{Yunhui Xie}.} \bibinfo{year}{2024}\natexlab{}.
\newblock \showarticletitle{Can robots elicit empathy? The effects of social robots’ appearance on emotional contagion}.
\newblock \bibinfo{journal}{\emph{Computers in Human Behavior: Artificial Humans}} \bibinfo{volume}{2}, \bibinfo{number}{1} (\bibinfo{year}{2024}), \bibinfo{pages}{100049}.
\newblock


\bibitem[Zhang et~al\mbox{.}(2023)]%
        {zhang2023effects}
\bibfield{author}{\bibinfo{person}{Shengliang Zhang}, \bibinfo{person}{Guanyu Tang}, \bibinfo{person}{Xiaodong Li}, {and} \bibinfo{person}{Ai Ren}.} \bibinfo{year}{2023}\natexlab{}.
\newblock \showarticletitle{The effects of appearance personification of service robots on customer decision-making in the product recommendation context}.
\newblock \bibinfo{journal}{\emph{Industrial Management \& Data Systems}} \bibinfo{volume}{123}, \bibinfo{number}{2} (\bibinfo{year}{2023}), \bibinfo{pages}{578--595}.
\newblock


\bibitem[Zhou et~al\mbox{.}(2022)]%
        {zhou2022responsive}
\bibfield{author}{\bibinfo{person}{Mohan Zhou}, \bibinfo{person}{Yalong Bai}, \bibinfo{person}{Wei Zhang}, \bibinfo{person}{Ting Yao}, \bibinfo{person}{Tiejun Zhao}, {and} \bibinfo{person}{Tao Mei}.} \bibinfo{year}{2022}\natexlab{}.
\newblock \showarticletitle{Responsive listening head generation: a benchmark dataset and baseline}. In \bibinfo{booktitle}{\emph{European Conference on Computer Vision}}. Springer, \bibinfo{pages}{124--142}.
\newblock


\bibitem[Zhu et~al\mbox{.}(2023)]%
        {zhu2023key}
\bibfield{author}{\bibinfo{person}{Xiaoling Zhu}, \bibinfo{person}{Wenrui Liang}, \bibinfo{person}{Wenjun Xv}, {and} \bibinfo{person}{Yimin Wang}.} \bibinfo{year}{2023}\natexlab{}.
\newblock \showarticletitle{The Key Strategies for Increasing Users’ Intention of Self-Disclosure in Human-Robot Interaction through Robotic Appearance Design}. In \bibinfo{booktitle}{\emph{SHS Web of Conferences}}, Vol.~\bibinfo{volume}{165}. EDP Sciences, \bibinfo{pages}{01012}.
\newblock


\bibitem[Z{\l}otowski et~al\mbox{.}(2016)]%
        {zlotowski2016appearance}
\bibfield{author}{\bibinfo{person}{Jakub Z{\l}otowski}, \bibinfo{person}{Hidenobu Sumioka}, \bibinfo{person}{Shuichi Nishio}, \bibinfo{person}{Dylan~F Glas}, \bibinfo{person}{Christoph Bartneck}, {and} \bibinfo{person}{Hiroshi Ishiguro}.} \bibinfo{year}{2016}\natexlab{}.
\newblock \showarticletitle{Appearance of a robot affects the impact of its behaviour on perceived trustworthiness and empathy}.
\newblock \bibinfo{journal}{\emph{Paladyn, Journal of Behavioral Robotics}} \bibinfo{volume}{7}, \bibinfo{number}{1} (\bibinfo{year}{2016}), \bibinfo{pages}{000010151520160005}.
\newblock


\end{thebibliography}
\end{document}